\documentclass[runningheads]{llncs}
\usepackage{booktabs}  
\usepackage{verbatim}
\usepackage{graphicx}
\usepackage{caption}
\usepackage{hyperref}

\usepackage{nicefrac}
\usepackage{amsmath}
\usepackage{multirow}
\renewcommand{\bf}{\textbf}

\begin{document}
\title{Reg R-CNN: Lesion Detection and Grading under Noisy Labels}
\titlerunning{}
%
\author{Gregor N. Ramien \and Paul F. Jaeger\and Simon A. A. Kohl\thanks{now with the Karlsruhe Institute of Technology and DeepMind (London)} \and Klaus H. Maier-Hein}
\authorrunning{Ramien et al.}
\institute{Division of Medical Image Computing, German Cancer Research Center (DKFZ), Heidelberg, Germany}
%
\maketitle              
\begin{abstract}
For the task of concurrently detecting and categorizing objects, the medical imaging community commonly adopts methods developed on natural images. Current state-of-the-art object detectors are comprised of two stages: the first stage generates region proposals, the second stage subsequently categorizes them. Unlike in natural images, however, for anatomical structures of interest such as tumors, the appearance in the image (e.g., scale or intensity) links to a malignancy grade that lies on a continuous \textit{ordinal scale}. While classification models discard this ordinal relation between grades by discretizing the continuous scale to an unordered “bag of categories”, regression models are trained with distance metrics, which preserve the relation. This advantage becomes all the more important in the setting of label confusions on ambiguous data sets, which is the usual case with medical images. To this end, we propose Reg R-CNN, which replaces the second-stage classification model of a current object detector with a regression model. We show the superiority of our approach on a public data set with 1026 patients and a series of toy experiments.
Code will be available at \href{https://www.github.com/MIC-DKFZ/RegRCNN}{\texttt{github.com}\texttt{/MIC-DKFZ/}\texttt{RegRCNN}}.
\end{abstract}
\section{Introduction}
The task of concurrently detecting and categorizing objects has been extensively studied in classic computer vision \cite{maskrcnn,retina}. In medical image computing, numerous approaches have been proposed to predict lesion locations and gradings, most of them in a supervised manner utilizing manual annotations. However, when adopting state-of-the-art object detectors for end-to-end lesion grading, one has to account for an inherent difference in the data: The grading of lesions denotes a subjective discretization of naturally continuous and ordered features (such as scale or intensity) to semantic categories with clinical meaning (e.g., BI-RADS score, Gleason score \cite{nagpal:2018}, PI-RADS score, TNM staging). This is in contrast to typical tasks on natural images, where categories can be described as an unordered set (no natural ordinal relation exists between dogs and cars). Hence, current object detectors phrase the categorization as a classification task and are trained using the cross-entropy loss, not considering the continuous ordinal relation between classes (see Sec. \ref{sec:theory}).

In this paper, we account for the ordinal information in lesion appearance and derived categories, aiming to improve model performance. To this end, we propose Reg R-CNN, which replaces the classification model of Mask R-CNN \cite{maskrcnn}, a state-of-the-art object detector, with a regression model.
Regression models utilize distance metrics, i.e., models are trained directly on the underlying continuous scale, which has the following major benefit in the setting of lesion grading on medical images:

Medical data sets often exhibit high ambiguity that is reflected in the variability of the human annotations. Under the assumption that class confusions follow a distribution around the underlying ground truth, distance metrics used in regression such as the L1-distance are more tolerant to mild deviation from the target value as opposed to the categorical cross entropy which penalizes all off-target predictions in equal measure \cite{Ghosh:2017:robust}.

We empirically show the superiority of Reg R-CNN on a public data set with 1026 patients and a series of toy experiments with code made publicly available.
\section{Methods}
\subsection{Regression vs. Classification Training}
\label{sec:theory}
In order to see why we expect the training of regression models to be more robust to label noise than classification models for the case when target classes lie on a continuous scale, let us first revisit the objective commonly minimized by classifiers. This objective is the cross entropy (CE), defined as
\begin{equation}
    H\left(\textbf{p}, \textbf{q}; \textbf{X}\right) = - \sum_{j} p_j(\textbf{X}) \log q_j(\textbf{X})
\label{eq:ce_bare}
\end{equation}
between a target distribution $\textbf{p}(\textbf{X})$ over discrete labels $j \in C$ and the predicted distribution $\textbf{q}(\textbf{X})$ given data $\textbf{X}$. For mutually exclusive classes, the target distribution is given by a delta distribution $\textbf{p}(\textbf{X})=\{\delta_{ij}\}_{j\in C}$. 

To produce a prediction $\textbf{q}(\textbf{X})$, the network's logits $\textbf{z}(\textbf{X})$ are squashed by means of a softmax function:
\begin{equation}
    \textbf{q}(\textbf{X}) = \frac{e^{\textbf{z}(\textbf{X})}}{\sum_{k\in C} e^{z_k(\textbf{X})}}, 
\end{equation}
which, plugged into Eq. \ref{eq:ce_bare} and given the target class $i$, leads to the loss term
\begin{equation}
       H = {\cal L}_{CE}(\textbf{p} = \delta_{ij}, \textbf{q};\textbf{X}) =  -z_i + log \textstyle \sum_k e^{z_k}.
\label{eq:ce}
\end{equation}

From Eq. \ref{eq:ce} it is apparent that the standard CE loss treats labels as an unordered bag of targets, where all off-target classes ($j \neq i$) are penalized in equal measure, regardless of their proximity to the target class $i$. Distance metrics on the other hand, as their name suggests, take into account the distance of a prediction to the target. This lets the loss scale in the deviation of prediction to target. Allowing to be more accepting of mild discrepancies, it better accommodates for noise from potentially conflicting labels in settings where the target labels lie on a continuum.

In the range of experiments below, we compare classification against regression setups, for which we employed the smooth L1 loss \cite{Huber:1964} given by 
\begin{equation}
\label{eq:huber}
    \mathcal{L}_{reg}(p,t) = 
    \begin{cases} 
    \nicefrac{1}{2}(t-p)^2,  & |t-p|<1 \\
    |t-p|-\nicefrac{1}{2},  & \text{otherwise} 
    \end{cases}
\end{equation}
for predicted value $p$ and target value $t$. Other works have investigated adaptions to the CE loss to account for noisy labels in classification tasks, e.g.  \cite{Tanno:2019,Zhang:2018}. Our approach is complementary to those works as it exploits label continua on medical images.

\subsection{Reg R-CNN \& Baseline}
\begin{figure}[!t]
    \centering
    \includegraphics[width=\textwidth]{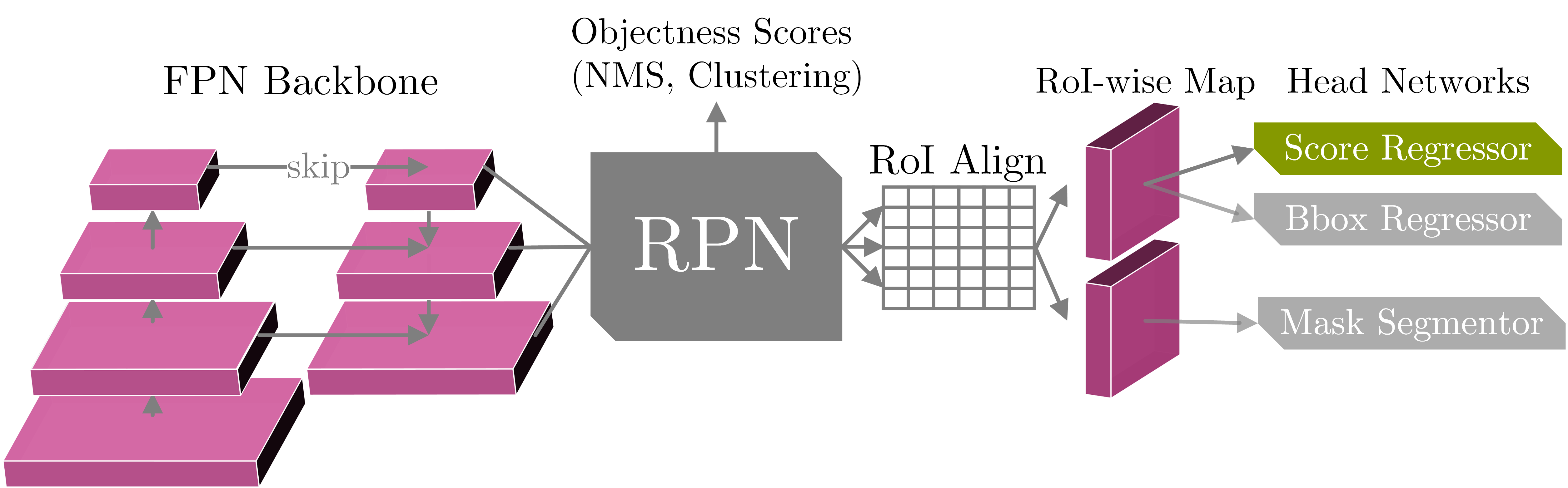}
    \caption{Reg R-CNN for joint detection and grading of objects. The architecture is closely related to Mask R-CNN \cite{maskrcnn}, where grading is done with a classification head instead of the displayed ``Score Regressor" head network. FPN denotes the feature pyramid network \cite{fpn}, RPN denotes the region proposal network and RoIAlign is the operation which resamples object proposals to a fixed-sized grid before categorization.}
    \label{fig:architecture}
    \vspace{-0mm}
\end{figure}

The proposed Reg R-CNN architecture is based on Mask R-CNN \cite{maskrcnn}, a state-of-the-art two-stage detector. In Mask R-CNN, first, objects are discriminated from background irrespective of class, accompanied by bounding-box regression to generate region proposals of variable sizes. Second, proposals are resampled to a fixed-sized grid and fed through three head networks: A classifier for categorization, a second bounding-box regressor for refinement of coordinates, and a fully convolutional head producing output segmentations (the latter are not further used in this study except for the additional pixel-wise loss during training). Reg R-CNN (see Fig. \ref{fig:architecture}) simply replaces the classification head by a regression head, which is trained with the smooth L1 loss instead of the cross-entropy loss (see Sec. \ref{sec:theory}).

For the final filtering of output predictions, non-maximum suppression (NMS) is performed based on detection-confidence scores. In Mask R-CNN, these are provided by the classification head. Since the regression head does not produce confidences, we use the objectness scores from the first stage instead. 

In this study, we compare Reg R-CNN against Mask R-CNN as the classification counterpart of our approach. Only minor changes are made with respect to the original publication \cite{maskrcnn}: The number of feature maps in the region proposal network is lowered to 64 to account for GPU memory constraints. The poolsize of 3D RoIAlign (a 3D re-implementation of the resampling method used to create fixed-sized proposals) is set to (7, 7, 3) for the classification head and (14, 14, 5) for the mask head. The matching Intersection over Union (IoU) for positive proposals is lowered to 0.3. Objectness scores are used for the final NMS to reflect the desired disentanglement of detection and categorization tasks. 

Note that all changes apply to Reg R-CNN as well, such that the only difference between the models is the exchange of the classification head with a regression head.

\subsection{Evaluation}\label{sec:eval}%
Comparing the performance of regression to classification models requires taking into account additional considerations since both are trained along an upstream detection task.

In order to compare continuous regression and discrete classification outputs, we bin the continuous regression output after training, such that bin centers match the discrete classification targets.

What's more, the joint task of object detection and categorization is commonly evaluated using average precision (AP) \cite{Everingham:pascal2010}. However, AP requires per-category confidence scores, which are, as mentioned before, not provided by regression outputs. Instead, we borrow a metric commonly used in viewpoint estimation, the Average Viewpoint Precision (AVP) \cite{xiang:pascal3d}. Based on AVP, we phrase the lesion scoring as an additional task on top of foreground vs. background object detection: In order for a box prediction to be considered a true positive, it needs to match the ground-truth box with an IoU $>$ 0.1\footnote{This relatively low matching threshold respects the clinical need for coarse localization and exploits the non-overlapping nature of objects in 3D images.}, and additionally the malignancy prediction score is required to lie in the correct category bin. 
This way, AVP simultaneously measures both the detection and malignancy-scoring performance of the models. We additionally disentangle the task performances and separately report the AP of foreground vs. background detection (this poses an upper bound on AVP) and the bin accuracy. The latter is determined by selecting only true positive predictions according to the detection metric and counting malignancy-score matches with the target bin.
\begin{figure*}[!t]
\centering
\includegraphics[width=\textwidth]{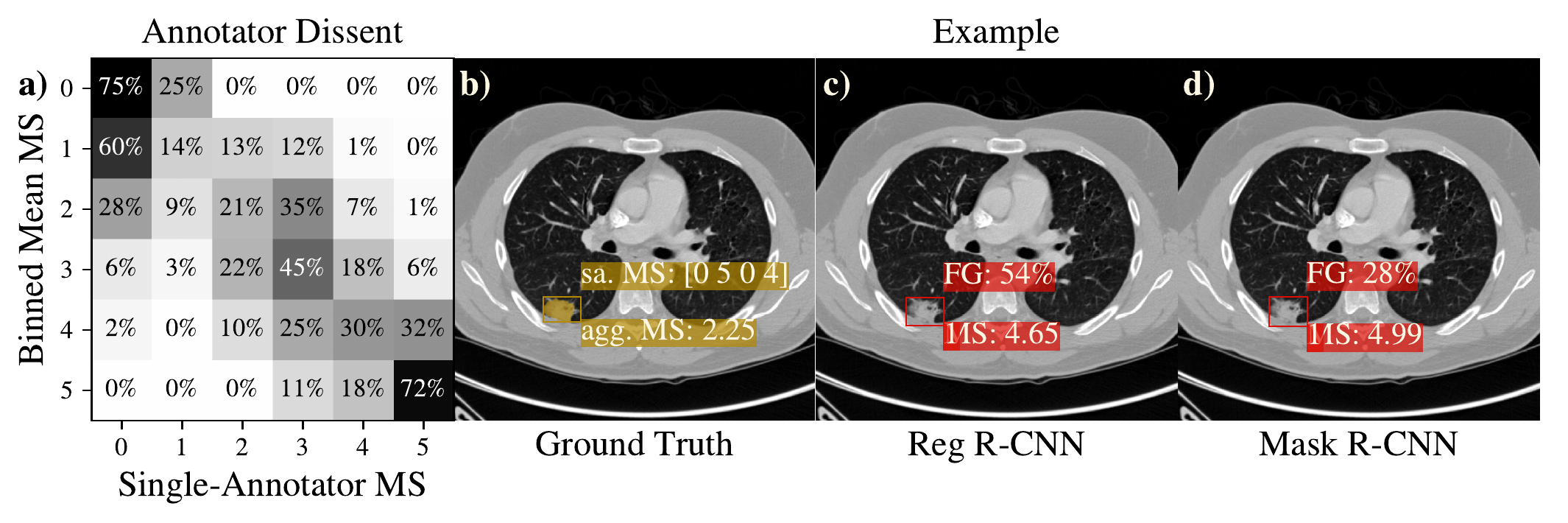}
\caption{\textbf{a)} A confusion-matrix-style display of annotator dissent in the LIDC data set. Rows represent the binned mean ratings of lesions (in place of the true class in a standard confusion matrix), columns the ratings of the corresponding single annotators. ``MS" means malignancy score. Matrix is row-wisely normalized, hence cell values indicate distribution of lesion ratings within a bin. \textbf{b)-d)} Example slice from the LIDC data set showing GT, Reg R-CNN, and Mask R-CNN prediction separately. GT note ``sa. MS" shows the single-annotator grades (grade 0 means no finding), ``agg. MS" their mean. In the predictions, ``FG" means foreground confidence (objectness score), ``MS" denotes the predicted malignancy score. Mask R-CNN MS can be non-integer due to Weighted Box Clustering \cite{paul:retinau}. Color symbolizes bin.}
\label{fig:lidc_dissent}
\vspace{-2mm}
\end{figure*}
\section{Experiments}
\subsection{Utilized data sets}
\subsubsection{Lung CT data set}
The utilized LIDC-IDRI data set consists of 1026 patients with annotations of four medical experts each \cite{lidc:data}. Having disposable multiple gradings from distinct annotators is a rare exception on medical images and allows to investigate the exhibited label noise \cite{kohl2018probabilistic}.

Full agreement, which we define as all raters assigning the same malignancy label to all lesions (RoIs) in a patient, is observed on a mere 163 patients (this includes patients void of findings by all raters). This corresponds to a rater disagreement with respect to the malignancy scoring on $84\%$ of the patients. On a lesion level (RoI-wise), the data set comprises 2631 lesions when considering all lesions with a positive label by at least one rater. This number drops to 1834, 1333, or 821, when requiring 2, 3, or 4 positive labels respectively. This shows that this data set's labelling is both ambiguous with respect to whether or not a lesion is present as well as the prospective lesion's grading. The first ambiguity type has bearings on the detection head's performance, while the second type influences the network's classification or, respectively, regression head.

In order to evaluate the grading performance, the following malignancy statistics include only patients with at least one finding. Among those, we count 99 lesions ($3.8\%$ of all lesions) with full rater agreement, leaving disagreement on 2532 (or $96.2\%$). The standard deviation of the 4 graders averaged over all lesions amounts to $1.05$ malignancy-score values (ms). In Fig. \ref{fig:lidc_dissent} a), we show how the single graders' malignancy ratings differ given the binned mean rating. The figure reveals significant label confusion across adjacent labels and even beyond.
Figs. \ref{fig:lidc_dissent} b)-d) display example Reg and Mask R-CNN predictions next to the corresponding ground truth.

In order to investigate the models' performance under label noise, we randomly sample a malignancy score (MS) for a given lesion from the 4 given gradings at each training iteration. At test time, we however employ the lesions' mean malignancy score as the ground truth label, which allows to evaluate against a ground truth of reduced noise.
\begin{figure}[h]
    \centering
    \vspace{-5mm}
    \includegraphics[width=\textwidth]{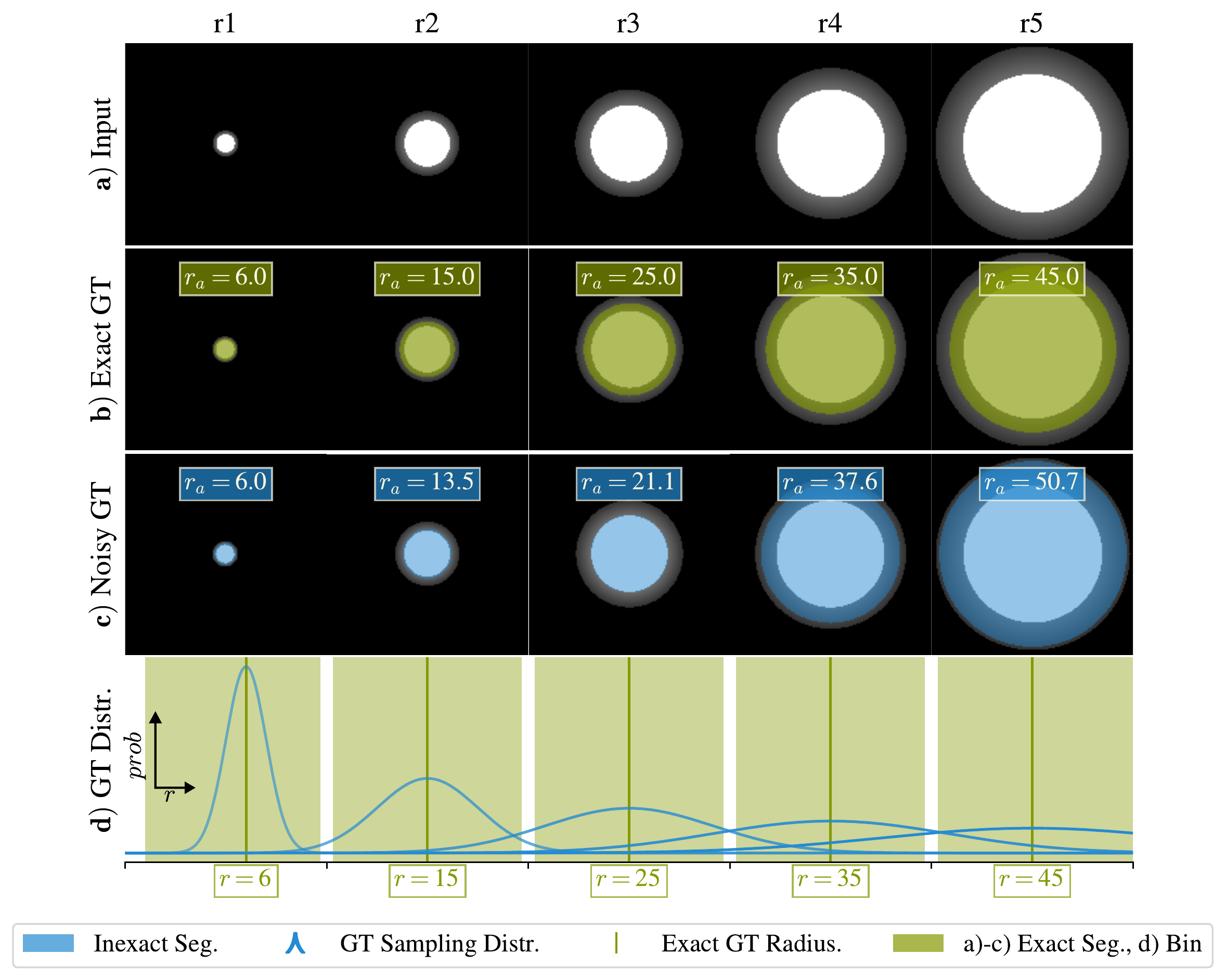}
    \caption{\textbf{a)} Cylinders (2D projections) of all five categories (r1-r5) in the toy experiment. \textbf{b)} Exact GT. \textbf{c)} Examples of a noisy GT for each category (r1-r5). $r_a$ indicates the annotated radius (target regression value). \textbf{d)} Gaussian sampling distributions used to generate the noisy GT. Green vertical lines depict the exact ground-truth values, while blue lines are the corresponding label-noise distributions. Green rectangles are the bins (borders enlarged for illustration) used for training of the classifier as well as for evaluation of both methods. Note that distributions reach into neighboring bins leading to label confusions.}
    \label{fig:toy_sketch}
    \vspace{-6mm}
\end{figure}
\subsubsection{Toy data set}
To analyze the performance of Reg R-CNN vs. Mask R-CNN on an artificial data set with label noise on a continuous scale, we designed a set of 3D toy images. The associated task is the joint detection and categorization of cylinders, where five categories are distinguished as cylinders of five different radii. In order to simulate label confusion, Gaussian noise is added to the isotropic target radii during training, sampled with standard deviation $\sigma=\nicefrac{r}{6}$ around object radius $r$, as depicted in Figs. \ref{fig:toy_sketch} c) and d). This causes targets (especially of large-radius objects) to be shifted into wrong, yet mostly adjacent target bins. Fig. \ref{fig:toy_sketch} a) portrays that these ambiguities are imprinted on the images as a belt of reduced intensity with width $2\sigma$ around the actual radius. At test time, model predictions are evaluated against the exact target radii without noise. The data set consists of 1.5k randomly generated samples for training and validation, as well as a hold-out test set of 1k images. 
\subsection{Training \& Inference Setup}
Both the LIDC and the toy data set consist of volumetric images. In this study, we evaluate models both in 3D as well as 2D (slice-wise processing). For the sake of comparability, all methods are implemented in a single framework and run with identical hyperparameters.
Networks are trained on patch crops of sizes $160\times160\times96$ (LIDC) and $320\times320\times8$ (toy), oversampling of foreground regions is applied. Class imbalances in object-level classification losses are accounted for by stochastically mining the hardest negative object candidates according to softmax probability. 

On LIDC, models are trained for 130 epochs, each composed of 200 batches with size 8 (20) in 3D (2D) using the Adam optimizer \cite{adam} with default settings at a learning rate of $10^{-4}$. Training is performed as a five-fold cross validation (splits: train $60\%$ / val $20\%$ / test $20\%$). At test time, we ensemble the four best performing models according to validation metrics over four test-time views (three mirroring augmentations) in each fold. Aggregation of box predictions from ensemble members is done via clustering and weighted averaging of scores and coordinates. Predictions from 2D models are consolidated along the z-axis by means of an adaption of NMS and evaluated against the 3D ground truth.
\subsection{Results \& Discussion}
Results are shown in Table \ref{tab:results}. In addition to the fold means of the metrics, we report the corresponding standard deviations. On LIDC, Reg R-CNN outperforms Mask R-CNN on both input dimensions and all three considered metrics. On the toy data set, Reg R-CNN shows superior performance in $\mathrm{AVP_{10}}$ and Bin Accuracy. $\mathrm{AP_{10}}$ reaches $100\%$ in both models indicating that the detection task is solved entirely, i.e., the object grading task has been isolated successfully (hence, results for $\mathrm{AVP_{10}}$ converge towards the Bin Accuracy). All experiments demonstrate the superiority of distance losses in the supervision of models performing continuous and ordered grading under noisy labels. Interestingly, there is a marked increase in performance for both setups when running in 3D as opposed to 2D, suggesting that additional 3D context is generally beneficial for the task.
\begin{table}[!t]
\setlength\tabcolsep{4pt}
  \caption{Results for LIDC and the toy data set. $\mathrm{AVP_{10}}$ measures joint detection and categorization performance, while $\mathrm{AP_{10}}$ measures the disentangled detection performance and Bin Accuracy shows categorization performance (conditioned on detection, see Sec. \ref{sec:eval})}
  \label{tab:results}
  \vspace{-2mm}
\setlength{\tabcolsep}{7pt}
\def\arraystretch{1.20}
\begin{center}
\begin{tabular}{lll|ccc}
& Dim & Network Head      & $\text{AVP}_{10}$ & $\text{AP}_{10}$       & Bin Accuracy  \\
\toprule
\multirow{4}{2mm}{\rotatebox[origin=c]{90}{LIDC}} 
&\multirow{2}{*}{3D}
& Reg R-CNN &\bf{0.259$\pm$0.035} &\bf{0.628$\pm$0.038} & \bf{0.477$\pm$0.035}\\
&& Mask R-CNN    & 0.235$\pm$0.027     &  0.622$\pm$0.029    & 0.411$\pm$0.026  \\
\cline{2-6}
&\multirow{2}{*}{2D}
&  Reg R-CNN &\bf{0.148$\pm$0.046}&\bf{0.414$\pm$0.052}&\bf{0.468$\pm$0.057} \\
&& Mask R-CNN    & 0.127$\pm$0.034     & 0.406$\pm$0.040     & 0.447$\pm$0.018    \\
\midrule
\multirow{4}{*}{\rotatebox[origin=c]{90}{Toy}}
&\multirow{2}{*}{3D}  
&  Reg R-CNN & \bf{0.881$\pm$0.014}         & 0.998$\pm$0.004	& \bf{0.887$\pm$0.014} \\
&& Mask R-CNN    & 0.822$\pm$0.070         & \bf{1.000$\pm$.000} & 0.826$\pm$0.069 \\
\cline{2-6}
&\multirow{2}{*}{2D} 
&  Reg R-CNN & \bf{0.859$\pm$0.021}    & \bf{1.000$\pm$0.000}    & \bf{0.860$\pm$0.021} \\
&& Mask R-CNN    & 0.748$\pm$0.022	        & \bf{1.000$\pm$0.000}	& 0.748$\pm$0.021 \\
\end{tabular}
\end{center}
\vspace{-8mm}
\end{table}
\section{Conclusion}
Simultaneously detecting and grading objects is a common and clinically highly relevant task in medical image analysis. As opposed to natural images, where object categories are mostly well defined, the categorizations of interest for clinically relevant findings commonly leave room for interpretation. This ambiguity can bear on machine-learning models in the form of noisy labels, which may hamper the performance of classification models. Clinical label categories however often reside on a continuous and ordered scale, suggesting that label confusions are likely more frequent between adjacent categories. 

For this case, we show that both the performance of lesion detection and malignancy grading can be improved upon over a state-of-the-art detection model when simply trading its classification for a regression head and altering the loss accordingly. We document the success of the ensuing model Reg R-CNN on a large lung CT data set and on a toy data set that induces artificial ambiguity. We attribute the edge in performance to the loss formulation of the regression task, which naturally accounts for the continuous relation between labels and is therefore less prone to suffer from conflicting gradients from noisy labels.
\section{Outlook}
As Eq. \ref{eq:huber} shows, we employ a metric approach to ordinal data. In general, this is not hazard-free as model performance may suffer from the imposed metric if the scale actually is non-metric \cite{Liddell:2018}. In other words, our approach implicitly assumes the grading scale has sufficiently metric-like properties. To address this limitation, we plan to study alternative non-metric approaches in future work \cite{Feindt:2004:neurobayes,McCullagh:1980}.
\bibliography{UNSURE_9.bib}
\bibliographystyle{abbrv}
\end{document}